\documentclass[letterpaper, 10 pt, conference]{ieeeconf}  

\IEEEoverridecommandlockouts                              

\usepackage[percent]{overpic}
\usepackage{graphicx} 
\usepackage{epsfig} 
\usepackage{amsmath} 
\usepackage{color}
\usepackage{soul}
\usepackage{url}
\usepackage[textstyle]{SIunits}

\title{\LARGE \bf Longitudinal Dynamics Model Identification of an Electric Car \newline Based on Real Response Approximation}

\author{Salvador Dominguez$^{1}$, Ga\"{e}tan Garcia$^{1}$, Arnaud Hamon$^{1}$ and Vincent Fr\'{e}mont$^{1}$ 
\thanks{$^{1}$ These authors are with LS2N, Laboratoire des Sciences du Num\'{e}rique de Nantes, Centrale Nantes and CNRS, 1 rue de la No\"{e}, 44321 Nantes, France}%
}

\begin{document}

\maketitle
\thispagestyle{empty}
\pagestyle{empty}

\begin{abstract}

Obtaining a realistic and accurate model of the longitudinal dynamics is key for a good speed control of a self-driving car. It is also useful to simulate the longitudinal behavior of the vehicle with high fidelity. In this paper, a straightforward and generic method for obtaining the friction, braking and propulsion forces as a function of speed, throttle input and brake input is proposed. Experimental data is recorded during tests over the full speed range to estimate the forces, to which the corresponding curves are adjusted. A simple and direct balance of forces in the direction tangent to the ground is used to obtain an estimation of the real forces involved. Then a model composed of approximate spline curves that fit the results is proposed. Using splines to model the dynamic response has the advantage of being quick and accurate, avoiding the complexity of parameter identification and tuning of non-linear responses embedding the internal functionalities of the car, like \textit{ABS} or \textit{regenerative brake}. This methodology has been applied to \textit{LS2N's electric Renault Zoe} but can be applied to any other electric car. As shown in the experimental section, a comparison between the estimated acceleration of the car using the model and the real one over a wide range of speeds along a trip of about \unit{10}{\kilo\meter\per\hour} reveals only \unit{0.35}{\metrepersquaresecond} of error standard deviation in a range of $\boldsymbol{\pm{2}}$\metrepersquaresecond which is very encouraging.

\end {abstract}


\section{Introduction}
  In order to obtain an optimal response during speed control, it is of great importance to use an accurate model of the dynamics of the car. For normal and comfortable autonomous driving where tire efforts are under the slip limits, the longitudinal dynamics can be considered as decoupled from lateral dynamics, so that they can be modeled independently as a separate problem. This paper deals with obtaining a model of the longitudinal dynamics that can be used for speed control but also for simulation purposes in order to obtain a very similar response in simulation than the car in real conditions.
  The methodology proposed in this paper is based on the balance of forces in the direction  tangent to the ground, as well as the use of interpolating splines \cite{Mckinley1998} as models for fitting the response to the experimental data. In order to validate the models for motor propulsion and brake, we compare the estimated acceleration using the model with the real acceleration of the car during a test that covers a wide range of speeds and accelerations.  
  The structure of the paper is as follows: some state of the art related with longitudinal dynamics identification and modeling is presented in \textit{section II}. Then \textit{Section III} presents the equipment used, including the vehicle and the \textit{Drive by wire} system installed. In \textit{section IV} we present the methodology applied to obtain the real forces and the spline-based models for friction, propulsion and braking forces, which are put under test in \textit{section V}, where model-estimated and real accelerations are compared. Finally the conclusions are presented in \textit{Section VI}.

 \begin{figure}[!t]
    \centering
    \includegraphics[width=0.8\linewidth]{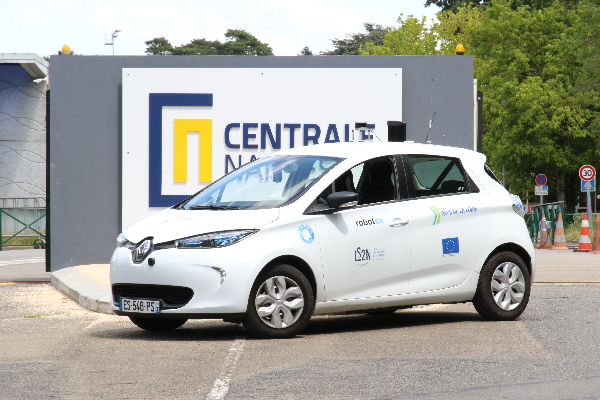}
    \caption{Autonomous \textbf{Renault ZOE ZE} used on the experiments}
    \label{fig:Vehicle}
 \end{figure}

\section{Related work}
 A model of the dynamics of the vehicle can be necessary for different purposes like for improving the control \cite{Chebly2017} \cite{Menhour2014}\cite{Attia2014}\cite{Majdoub2012}\cite{Dias2019}, or for simulating the behavior of the real vehicle like in \cite{Gomez2012} and \cite{Short2004} where the models are used in driving simulators. In the literature there exist many different ways to address the problem of modelling the dynamics of a vehicle. Some of them consider coupled longitudinal and lateral dynamics \cite{Chebly2017}\cite{Menhour2014}\cite{Macek2007}, but the most classical way to face this problem is to consider decoupled dynamics with complex parametric models. Generally, some assumptions are applied in order to reduce the number of parameters to a small set and simplify the process of identification, like in \cite{Attia2014}\cite{Short2004}\cite{James2018}. For instance, in \cite{James2018} the authors propose a simplified longitudinal model. The friction is modelled by a second order polynomial with two parameters and two terms, static and dynamic respectively. The propulsion force is considered proportional to gear-shaft torque, and finally the braking force is modelled as proportional to the brake pressure, both gear-shaft torque and brake pressure are obtained by reading the CAN bus of the car. These parameters are then identified in continuous-time using \textit{Prediction Error Method}. According to the authors, the resulting model represents the longitudinal dynamic behavior along the whole speed range. A more detailed parametric model is proposed in \cite{Majdoub2012}, where all the forces that intervene in longitudinal motion are explained and characterized proposing two theoretical control models in the state space, one for accelerating and the other for decelerating. However, the paper does not show how the numerical values for the parameters are obtained and does not consider throttle and brake as the control variables, which would be more practical for a generic vehicle. In \cite{Dias2019} the authors propose a discrete-time parametric model for engine drive and braking system that represents the behavior around the operation point in terms of speed and gear shift position, and the parameters are obtained by least squares method over pairs of empirical input and output data. The model is intended for low speed (under \unit{40}{\kilo\meter\per\hour}) and represents the relations throttle vs. speed and brake vs. speed on a flat surface, meaning that the model doesn't take into account the effect of gravity. Moreover, the friction force is somehow absorbed in the model. As far as we know, there are no references in the literature that address the problem of longitudinal dynamics modeling using interpolation curves. In our solution, we separately model using this technique the friction force versus speed, propulsion force versus throttle and speed and braking force versus brake strength and speed. As we are modeling forces, it is possible to compute the balance of all relevant forces in order to estimate the acceleration and conversely, for a desired acceleration to compute the balance of forces in order to get the necessary propulsion or braking forces and from them obtaining the necessary throttle or brake signal, using the corresponding inverse model. As we show in Section V, the results obtained are quite encouraging.

\section{The equipment}
\begin{figure}[!t]
    \centering
    \includegraphics[width=1.0\linewidth]{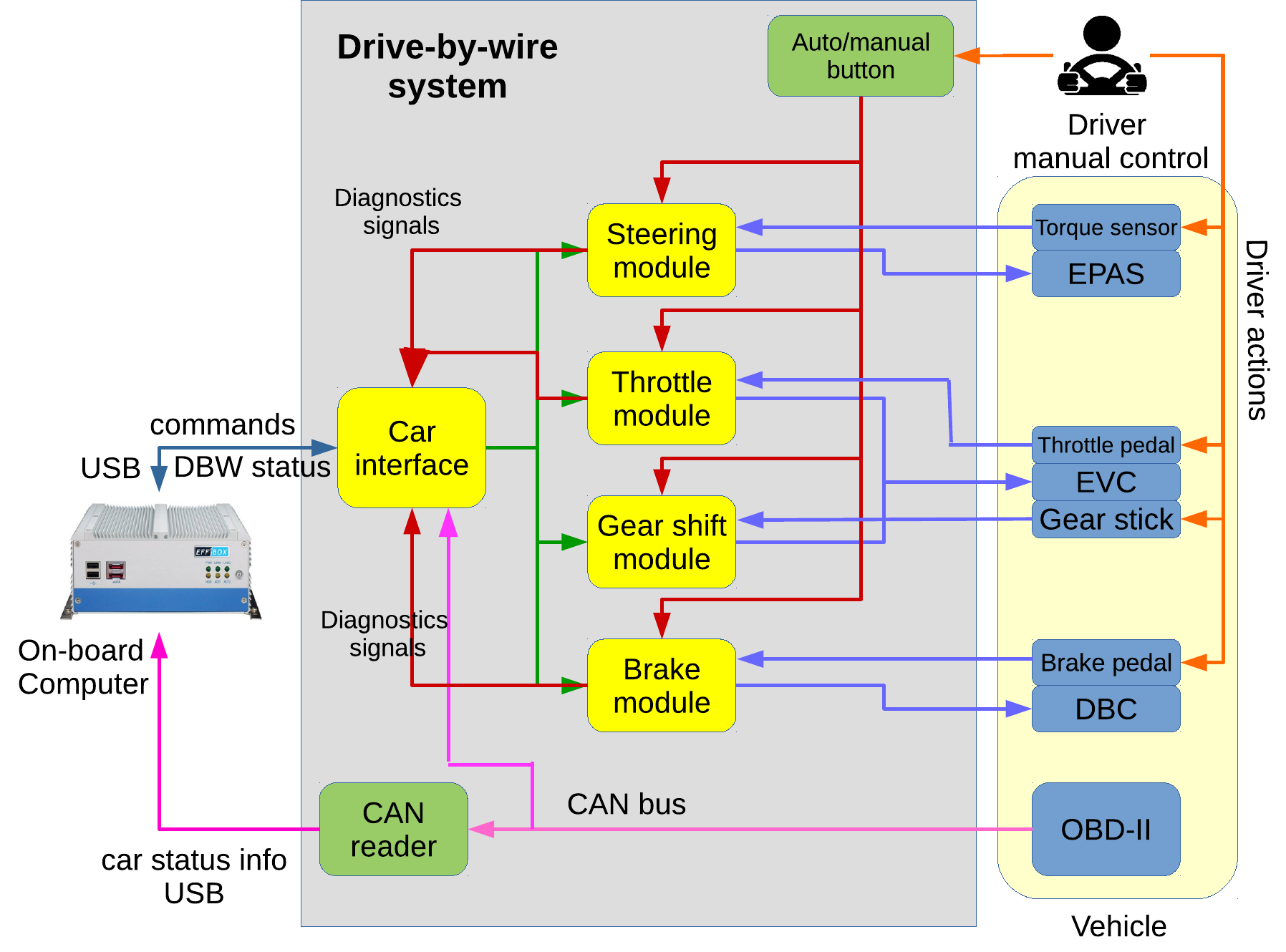}
    \caption{Block diagram of the drive-by wire system.}
    \label{fig:DriveByWire}
 \end{figure}

 The vehicle used for the experiments is an electric \textit{Renault ZOE ZE} of 2016 (see Fig.\ref{fig:Vehicle}) which has been converted into drive-by-wire, hereafter \textit{DBW}, using a system developed at \textit{LS2N} for which more information is available at \cite{LS2Nkit}. The system allows to control the steering, throttle, brake and gear shift \textit{Reverse-Neutral-Drive} position. Fig.~\ref{fig:DriveByWire} shows a schematic of the \textit{DBW} system. Every small block represents a hardware element and the arrows represent the interconnections between them, which are either digital or analog. The central block in pink represents the whole \textit{LS2N's DBW} kit. It is composed of a central module called \textit{Car Interface} from now on \textit{CI}, which interconnects the on-board computer with the rest of the \textit{DBW} architecture. This module is in charge, among other things, of the low level control of the steering angle and the brake strength applied. It also implements the safety rules that either stop the car or switch it to manual mode depending on the situation. The \textit{CI} is also in charge of the communication between the computer and the \textit{DBW}. It receives commands from the computer, \textit{e.g.} to set a specific steering angle, brake force or throttle value, and periodically sends the \textit{DBW} status information to the on-board computer. The on-board computer gets the current status of the car directly from the \textit{OBD-II} diagnostics connector through a \textit{CAN reader}. The CAN messages are decoded and important information is extracted from those messages, \textit{e.g.} the steering angle and angular velocity, the speed of each wheel, the position of the \textit{P-R-N-D} gear stick, and more. The steering, throttle, brake and gear shift modules are signal adaptor circuits that allow converting the control signals generated by the \textit{CI} module into the format used by Renault in this vehicle. So from the point of view of the car calculators, there is no difference whether the signals come from the actions performed by a person who drives the car in manual mode or by the \textit{DBW} kit in automatic mode. The block on the right side of the schematic represents the car itself and the different calculators integrated in the vehicle by the manufacturer.
 The \textit{CI} has two operation modes:
 
 \begin{itemize}
  \item \textbf{Direct mode}: This mode is intended for development purposes. It is useful for identifying the model of an specific actuator (throttle, brake or steering) by setting its input manually and analyzing the response.
  \item \textbf{Normal mode}: This is the mode in which the \textit{DBW} system normally works in autonomous navigation. It performs different low-level control loops to reach a specific steering angle or brake strength as reference. For practical reasons, the low-level control of the speed by manipulating the brake and throttle signals is performed by the on-board computer as it is for development purposes for testing different techniques of speed control.
\end{itemize}
  The tests to obtain the models are performed in \textit{Direct mode} as, in this mode, we can set fixed values of the pedal signals directly from computer. 
 
\section{Methodology}

 \begin{figure}[!t]
    \centering
    \includegraphics[width=0.6\linewidth]{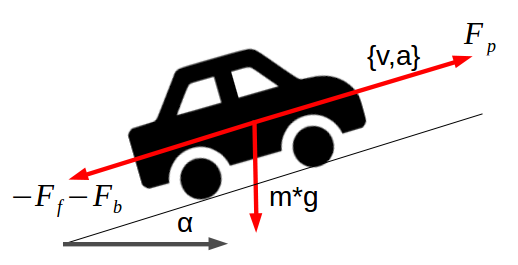}
    \caption{Relevant forces considered for longitudinal dynamics modeling. Eq.~(\ref{eq1}) is obtained by projecting the forces in the direction parallel to the ground.}
    \label{fig:LDMFigure}
 \end{figure}
 
  In this paper we propose to model the response of the car by applying a simple formula which is the balance of five forces in the direction tangent to the ground. The forces are: propulsive force of the motor $F_p$, ground-tangent weight component $mg\sin{\alpha}$, friction force $F_f$, braking force $F_b$ and the inertial force $m_{eq}a$, see Fig.~\ref{fig:LDMFigure} and Eq.~(\ref{eq1}).
  
  \begin{flalign}\label{eq1}
F_{p}-mg\sin(\alpha)-F_{f}-F_{b}=m_{eq}a
  \end{flalign}
  
 \begin{itemize}
  \item $F_{p}$: propulsive force of the motor. This is a priori an unknown value but it is directly related with the motor torque, which in turn depends on the throttle signal applied and the RPM, so $F_p$ must depend somehow on the throttle value and the speed of the car. Its sign is positive when pushing forward and negative otherwise.
  \item $m$: total mass. This includes the mass of the car itself \unit{1480}{\kilogram} according to specifications) and the additional mass like equipment and people inside. We estimate that the additional mass during the experiments was about \unit{200}\kilogram, so the total mass was about \unit{1680}{\kilogram}.
  
  \item $m_{eq}$: equivalent mass. It is the mass without rotating bodies which has the same kinetic energy as the car. The kinetic energy of a wheel of mass $m_w$, radius $R$ and inertia $J$ at speed $v$ in pure rolling motion is:
  
  \begin{flalign}\label{wheel_equiv_mass}
  E=\frac{1}{2}m_w v^2+\frac{1}{2}Jw^2=\frac{1}{2}(m_w+\frac{J}{R^2})v^2
    \end{flalign}\
  
  Assuming the wheels are the main rotating bodies of the car, Eq. \ref{wheel_equiv_mass} means that the equivalent mass of the car is obtained by adding $\frac{J}{R^2}$ for each wheel. With $J=$\unit{0.86}{\kilogrampersquaremetre} and $R$=\unit{0.29}\meter, the additional mass per wheel is \unit{10}{\kilogram}, for a total equivalent mass for the car $m_{eq}=$\unit{1720}{\kilogram}.
  \item $g$: gravity acceleration (\unit{9.81}{\metrepersquaresecond}).
  \item $\alpha$: slope of the road in radians.
  \item $F_{f}$: friction force. This is the resulting addition of forces due to internal friction between the motor and the wheels, between the tires and the road surface (rolling resistance) \cite{Bakker1987} and also aerodynamics resistance \cite{Hucho1993}. This is always a positive value in Eq.~(\ref{eq1}).
  \item $F_{b}$: braking force. This is the addition of forces applied by the brake disks and the regenerative brake. The regenerative brake is the brake force due to the fact that, when the car decelerates with no throttle applied, some of the kinetic energy is employed to charge the main batteries of the car. Like in the case of $F_f$, this is also a positive value in Eq.~(\ref{eq1}).
  \item $a$: linear acceleration of the car.
\end{itemize}

In each experiment we record the values of a set of variables at a sample rate of about \unit{100}{\hertz}:
 \begin{itemize}
  \item \textbf{Time stamp} of the data.
  \item \textbf{Speed} of the car.
  \item \textbf{Brake strength} value. This is the value of brake read on the CAN bus of the car. It represents the amount of brake applied but has no know relation to the brake force in Newton at this point.
  \item \textbf{Throttle} value. Value of throttle read on the CAN bus.
  \item \textbf{Slope} of the road. 
 \end{itemize}
 
 The full modelling process is based on Eq.~(\ref{eq1}). The process involves three steps:
 \begin{enumerate}
     \item For given conditions (transported mass, rolling surface, tire pressure, no wind ...), friction forces are a function of speed. Determine the function using tests where propulsion and brake forces are zero.
     \item Knowing friction from step one, determine the propulsion force as a function of throttle signal and speed, using tests where the brake force is zero.
     \item Determine brake force as a function of brake signal and speed using tests where the propulsion force is known.
 \end{enumerate}
 The reason why brake force depends on speed is twofold: \textit{regenerative braking} is used to recharge the battery when zero brake signal is used and speed is above a certain threshold, and also because of the \textit{ABS} system.
 Because the friction function is used in steps two and three, the corresponding recordings must be performed in the same conditions as step one for the friction values to be valid.
 
 All calculations are performed off-line. The slope is estimated using an \textit{Attitude Heading Reference System} filter \cite{Mahony2008}, compensated for the disturbances induced by longitudinal and centripetal accelerations of the car. Acceleration estimates use a filtered centered estimator to avoid phase shift.
 
 In the sequel, contrary to known literature, we pay attention to modelling the forces at low speeds. The reason is that they are important for smooth control in maneuvering phases, like parking. It is especially important for a vehicle with an automatic gearbox, where the only way to control very low speeds is by applying a braking force.
 
 Finally, it should be noted that the approximation functions in this work have been fitted manually. The reason is that, in the very low speeds, there is hardly any data, as these phases are extremely short. In these regions, the curves shown should be considered as a proposed response shape. Of course, once the shape has been chosen and validated, the process could be made automatic without any problem, should we have to model a new car.
 
 
\subsection{Friction curve}

By setting $F_p =0$ and $F_b=0$ in Eq.~(\ref{eq1}) we obtain $F_f$ as: 
   \begin{flalign}\label{eq2}
   F_{f}=-mg\sin{\alpha}-m_{eq}a
  \end{flalign}
In Eq.~(\ref{eq2}) all terms on the right side are measured. 
The experiment consists in accelerating the car to its maximum speed (about \unit{125}{\kilo\meter\per\hour}) and then letting the car decelerate in \textbf{neutral} gear shift position until it stops, while applying zero brake signal. Thus, there is no mechanical contact between the motor and the traction system, so $F_p=0$ and neither regenerative braking nor disk braking is applied so $F_b=0$. Also, in order not to disturb the longitudinal dynamic behavior too much, it is preferable not to perform the test on a windy day, where the force induced by the wind can be significant.

The values of friction obtained represent the real friction under the specific conditions of the experiment, like road surface, wind, payload, tire pressure, and include aerodynamic forces.

\begin{figure}[!t]
    \centering
    \includegraphics[width=1.0\linewidth]{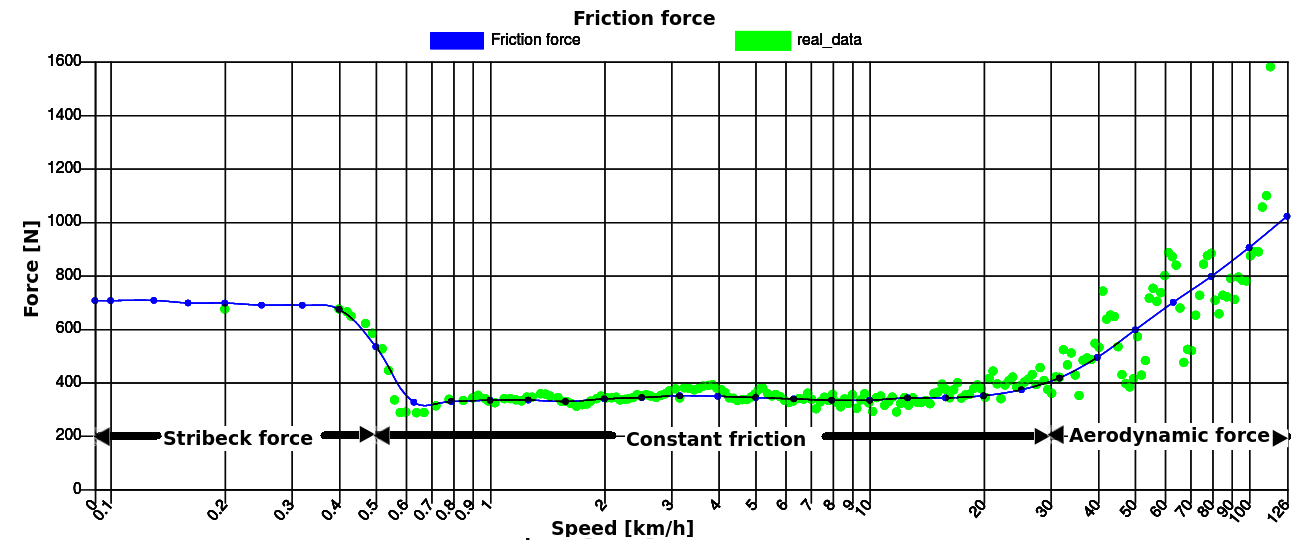}
    \caption{Friction force experimental results and corresponding proposed friction model}
    \label{fig:FrictionCurve}
 \end{figure}

Fig.~\ref{fig:FrictionCurve} shows the resulting friction curve, where points are the friction forces obtained using the data recorded, and the curve is the proposed continuous friction model. We use a logarithmic plot to properly visualize  the response at high, medium and low speeds. There are three different zones in the friction response:
 \begin{itemize}
  \item \textbf{Low speed range $\boldsymbol{(<}$\unit{0.5}{\kilo\meter\per\hour}$\boldsymbol{)}$}: In this range there is almost no experimental data but we observed that some points indicate higher friction than at medium speed. This may be due to the Stribeck force, which is the force that must be overcome in order to set the vehicle in motion, and is normally higher than the friction in the medium speed range.
  \item \textbf{Medium speed range $\boldsymbol{(}$\unit{0.5-30}{\kilo\meter\per\hour}$\boldsymbol{)}$}: In this range the results show a constant value of the friction force independently of the speed. This friction is due mainly to the rolling resistance between the tires and the road and must strongly depend on tire pressure and road surface characteristics.
  \item \textbf{High speed range $\boldsymbol{(>}$\unit{30}{\kilo\meter\per\hour}$\boldsymbol{)}$}: The aerodynamic resistance becomes significant. The friction force increases with speed. The linear shape in the logarithmic plot is consistent with aerodynamic forces proportional to the square of velocity.
 \end{itemize}

The friction curve can be used as a feed-forward estimation for speed control. Attention to lower speeds is justified by the fact that, with an automatic gearbox, the only way to control low speed is by applying a certain brake force. The model could be further improved by compensating it for changing conditions (added mass, changes in tire pressure...) but this is beyond the scope of this paper. 

\subsection{Motor propulsion force versus throttle and speed}
\begin{figure}[!t]
    \centering
    \includegraphics[width=1.0\linewidth]{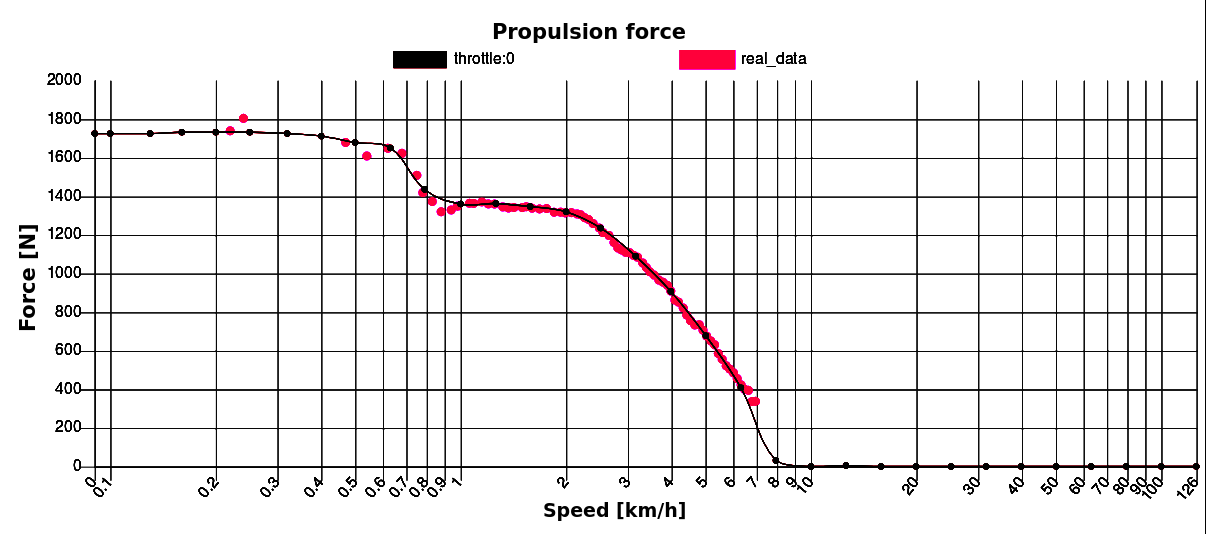}
    \caption{Experimental results of propulsion force for zero throttle value and corresponding proposed model. }
    \label{fig:PushForceThrottle0}
 \end{figure}

\begin{figure}[!t]
    \centering
    \includegraphics[width=1.0\linewidth]{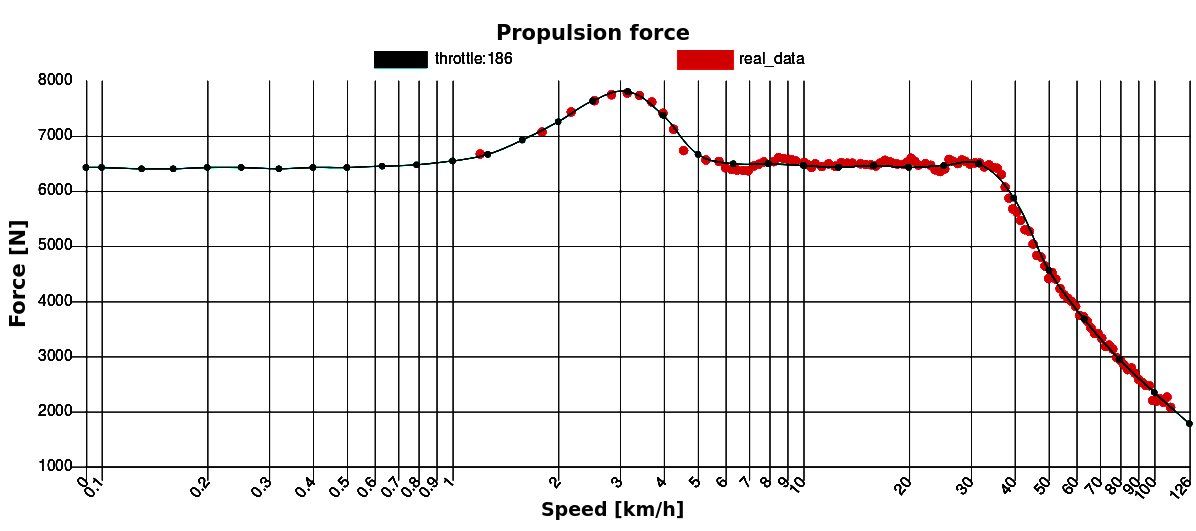}
    \caption{Experimental results of propulsion force for throttle value = 186 and corresponding proposed model.}
    \label{fig:PushForceThrottle186}
 \end{figure}

The propulsion force depends on the throttle applied and on the speed of the car. To obtain the corresponding curves we apply Eq.~(\ref{eq1}), using the friction curve obtained at the previous step. Of course, it is preferable to operate in similar environmental conditions (no wind, same road and tire pressure). In our case, the data was recorded in the same place and right after recording the friction data. The tests consist in letting the car to accelerate from zero speed until getting a steady speed while applying a constant throttle value. During the whole experiment, the acceleration is positive, so the braking force due to regenerative brake is zero, and we do not apply external brake, so we can consider that the braking force term in Eq.~(\ref{eq1}) is zero  ($F_b=0$). 
So $F_p$ is obtained as: 
  \begin{flalign}\label{eq4}
   F_{p}=F_f+mg\sin{\alpha}+m_{eq}a
  \end{flalign}
In this Eq. all the terms are known. As an example the results obtained for minimum ($throttle=0$) and maximum throttle applied ($throttle=186$) are shown in Fig.~\ref{fig:PushForceThrottle0} and \ref{fig:PushForceThrottle186}, overlaid with their respective proposed propulsion force model. As expected on a vehicle with automatic gear shift, a motor torque is applied to make the car move at low speed, in this case up to around \unit{8}{\kilo\meter\per\hour} on a horizontal flat surface, even with zero throttle signal. In the $throttle=0$ case there is no experimental data over \unit{7}{\kilo\meter\per\hour} because a steady speed is reached. The proposed model is based on the trend of the curve and on the fact that the propulsion force cannot be negative, as it would be a braking force. In the maximum throttle case, there is no information at the very low speed range \unit{[0-1]}{\kilo\meter\per\hour}, but we consider that for an electric motor there is no reason for the propulsion force to be different from what it is at \unit{1}{\kilo\meter\per\hour}, which is a very slow speed. So for the curves where we do not have information in this range we propose an almost constant value of $F_p$.
 
\begin{figure}[!t]
    \centering
    \includegraphics[width=1.0\linewidth]{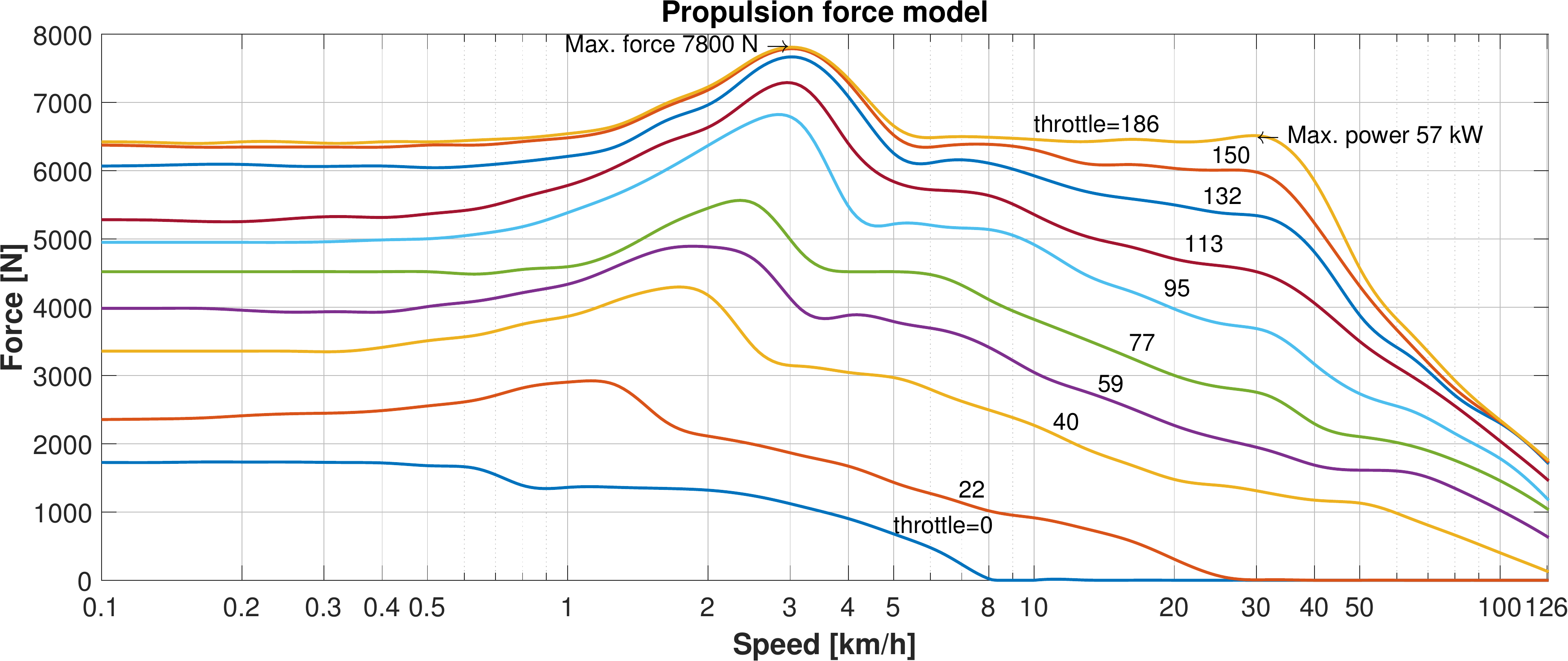}
    \caption{Proposed propulsion force model at different throttle values, taking into account the experimental data. Note that the maximum effective power obtained is about $57~kW$ and the maximum torque $7800~N$}
    \label{fig:PushForceModel}
 \end{figure}

 The proposed model for the propulsion force versus speed for different values of throttle is shown in Fig.~\ref{fig:PushForceModel}. Even though the results are shown as a set of curves, the actual model is a two dimensional surface where the axes are: Speed, throttle and propulsion force. Splines are defined in the speed-force planes as shown, and in throttle-force planes. Thus, a propulsion force can be calculated for any value of speed and throttle. Note that the propulsion force starts saturating over $throttle = 150$ and the optimal torque regime moves from \unit{1}{\kilo\meter\per\hour} to \unit{3}{\kilo\meter\per\hour} as the throttle value increases from 0 to the maximum value of 186. Specially on the curve corresponding to the maximum throttle value, it can be seen that there is a speed range where the propulsion force is more or less constant (\unit{6300}{\newton}) between $5$ and \unit{30}{\kilo\meter\per\hour}, then the torque decreases following a constant power curve of about \unit{57}{\kilo\watt}. According to manufacturer specifications, the nominal power of the car is higher (\unit{65}{\kilo\watt}). A possible reason could be that the maximum throttle signal generated by the DBW kit is lower than the maximum value obtained by fully pushing the throttle pedal.
 
\subsection{Braking force versus brake strength curves}
\begin{figure}[!t]
    \centering
    \includegraphics[width=1.0\linewidth]{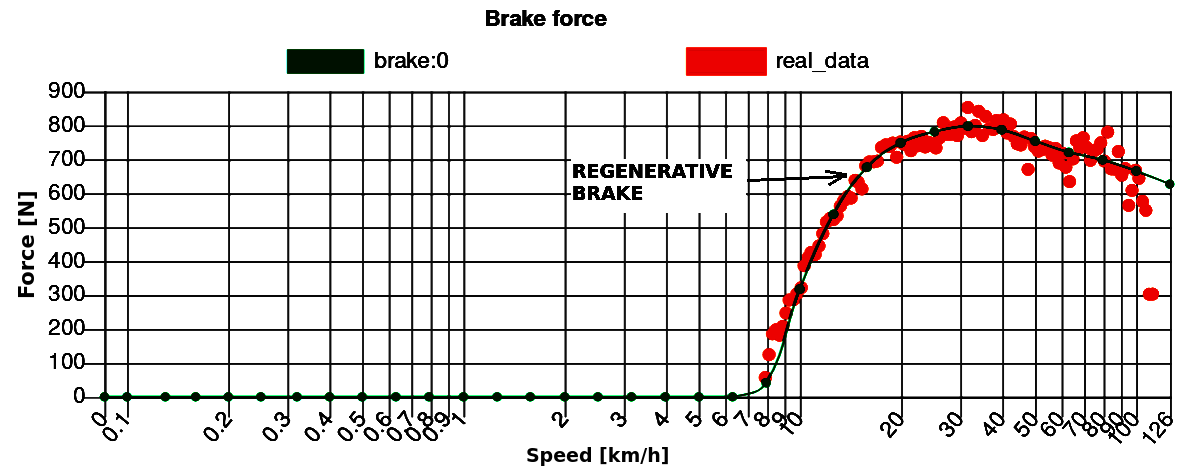}
    \caption{Experimental results of braking force for brake signal = 0 and corresponding proposed model.}
    \label{fig:BrakeForceBrake0}
 \end{figure}

\begin{figure}[!t]
    \centering
    \includegraphics[width=1.0\linewidth]{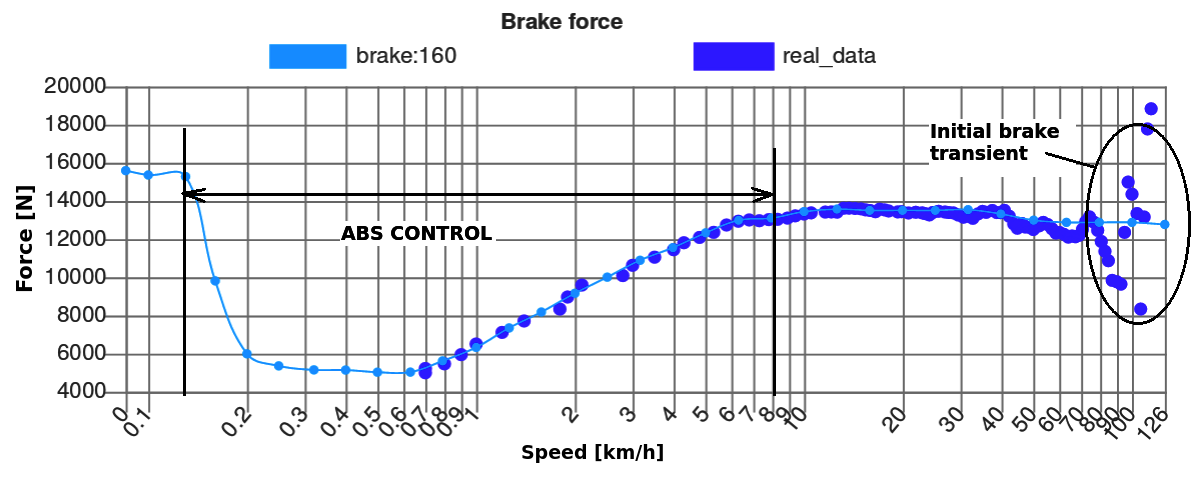}
    \caption{Experimental results of braking force for brake signal = 160 and corresponding model}
    \label{fig:BrakeForceBrake160}
 \end{figure}

The experiments to obtain the braking force model consist in starting from the highest speed of the car (about \unit{125}{\kilo\meter\per\hour}), and applying a constant brake signal until the car stops. The process is repeated for various values within the whole range of brake signals.
As for the friction and propulsion force we apply Eq.~(\ref{eq1}), taking into account that the friction curve is known, the gear shift is in \textbf{drive} position and that the throttle applied during the brake tests is zero, that is, we have to apply the curve of propulsion force corresponding to $throttle=0$ in Eq.~(\ref{eq1}), which is the curve shown in Fig.~\ref{fig:PushForceThrottle0}, thus obtaining Eq.~(\ref{eq5}). 
  \begin{flalign}\label{eq5}
   F_{b}=F_{p}(ttl=0)-F_f-mg\sin{\alpha}-m_{eq}a
  \end{flalign}
As an example the curves for ${brake=0}$ and ${brake=160}$ are shown in Fig.~\ref{fig:BrakeForceBrake0} and \ref{fig:BrakeForceBrake160}.
 
 In Fig.~\ref{fig:BrakeForceBrake0}, as no disk brake has been applied, the braking force must come from the regenerative braking system. According to the results obtained, the regenerative braking system is only effective over about \unit{8}{\kilo\meter\per\hour}. Under that speed there are no measures, as a steady speed is reached. The model proposed at the lower range of speeds is based on the trend of the experimental data and on the fact that the braking force cannot be negative as it would rather be a propulsion force. The regenerative braking model, which is the braking model with zero brake signal (Fig.~\ref{fig:BrakeForceBrake0}) must only be applied when there is no throttle signal since, when a non-zero throttle is applied the regenerative braking system is deactivated. In that case, if no brake is applied, the braking force is $F_b=0$.
 
\begin{figure}[!t]
    \centering
    \includegraphics[width=1.0\linewidth]{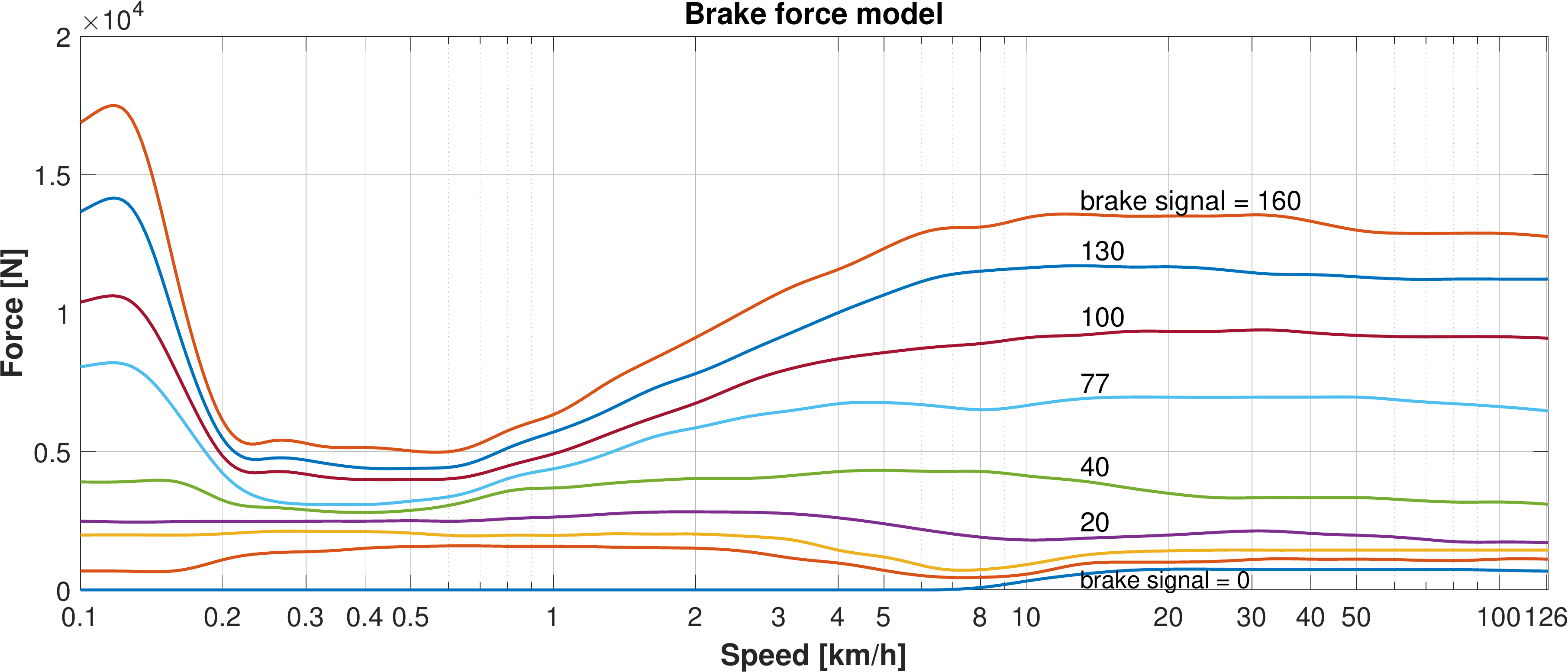}
    \caption{Brake model at different brake signal values.}
    \label{fig:BrakeForceModel}
 \end{figure}

 Fig.~\ref{fig:BrakeForceModel} shows the curves corresponding to the braking force obtained by applying Eq.~(\ref{eq5}) at different values of brake strength. Like in the case of the propulsion force, the actual model is a function of two variables defined by a set of splines in speed-force and brake signal-force planes. In figs.~\ref{fig:BrakeForceBrake160} and \ref{fig:BrakeForceModel}, we can see the effect of the \textit{Anti-Blocking-System (ABS)} in the low speed range. In order to prevent wheel slip, the \textit{ABS} reduces the brake force applied to the brake disks, the braking force being restored when the \textit{ABS} stops.
 
\begin{figure}[!t]
    \centering
    \includegraphics[width=1.0\linewidth]{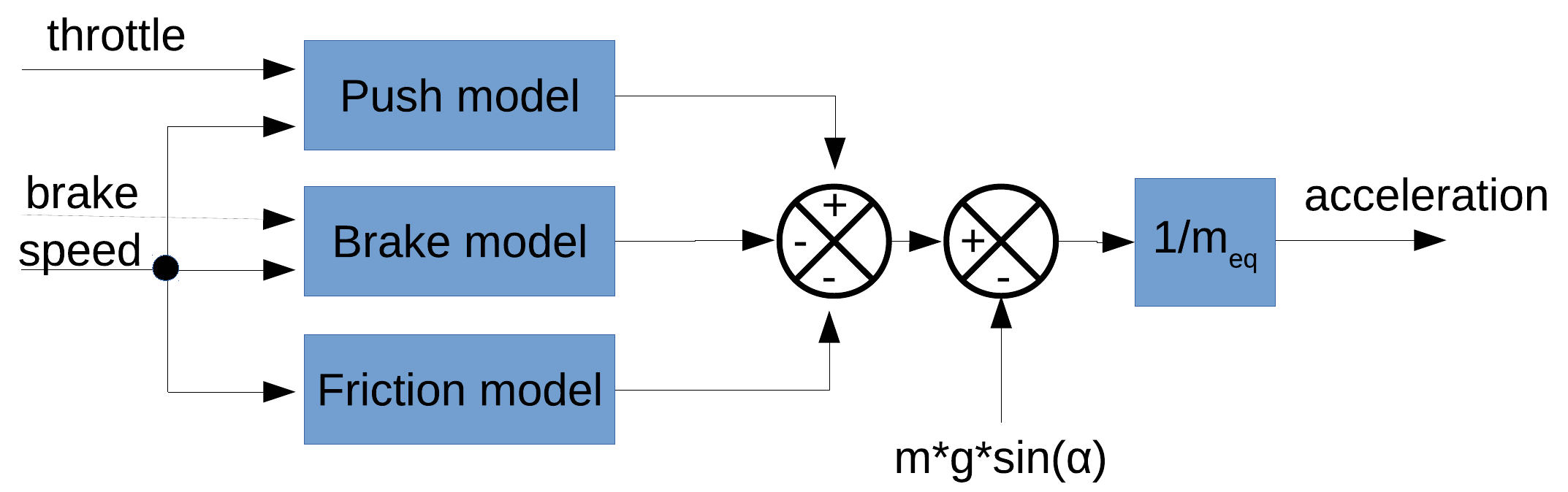}
    \caption{Using the models to estimate the forces in a simulation.}
    \label{fig:DirectModelSimulation}
 \end{figure}

\begin{figure}[!t]
    \centering
    \includegraphics[width=1.0\linewidth]{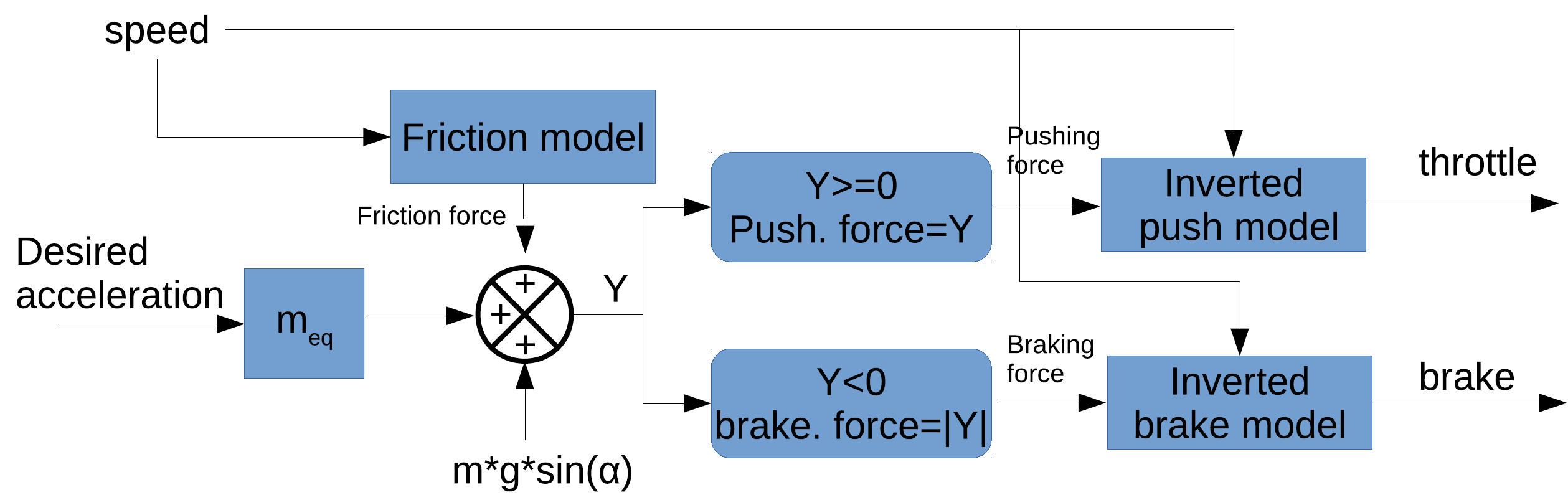}
    \caption{Using the inverted models to control the speed of the car.}
    \label{fig:InvertedModelControl}
 \end{figure}

 With this method, we build the models for friction, propulsion and brake, which provide an estimation of the forces. Thus, we can calculate the direct dynamic model of the car, suitable for simulation purposes and illustrated Fig.~\ref{fig:DirectModelSimulation}. It is also possible to invert the propulsion force model to find the throttle signal as a function of speed  and propulsion force, and do the same for the brake model, using new sets of splines. Then we can calculate the inverse dynamic model of the car, illustrated Fig.~\ref{fig:InvertedModelControl}. It is suitable for speed control, with the desired acceleration in Fig.~\ref{fig:InvertedModelControl} being typically calculated by the speed controller.
 

\section{Experimental results. Model validation}
In order to validate the models obtained, we have recorded the values of speed, slope, throttle and brake along a drive of approximately \unit{9.84}{\kilo\meter} in length. We use the recorded data to estimate the acceleration of the car (see Fig.~\ref{fig:DirectModelSimulation} and compare it to the measured acceleration. 

\begin{figure}[!t]
    \centering
    \includegraphics[width=1.0\linewidth]{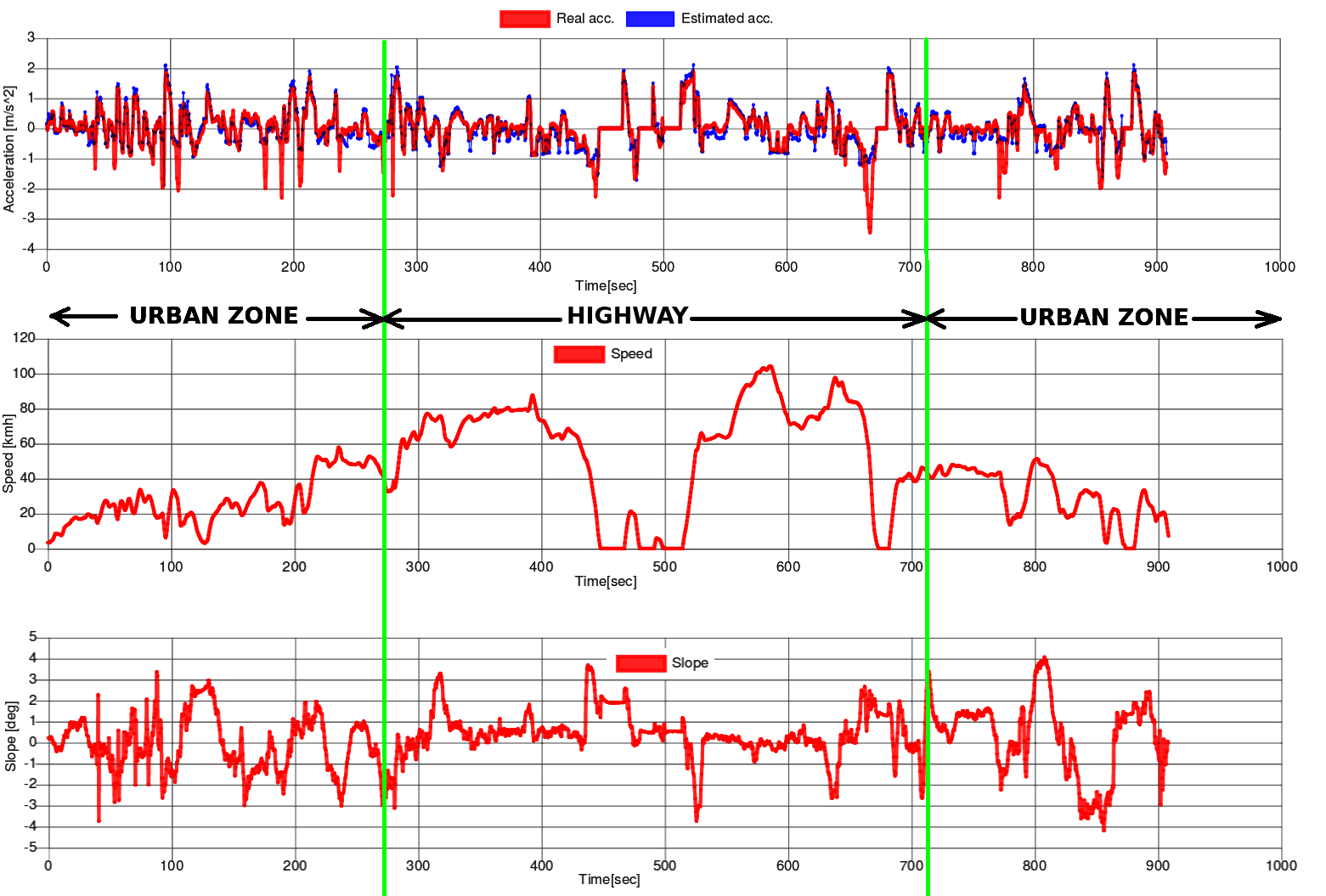}
    \caption{Time-line comparison of real and model-estimated accelerations (see~Fig.\ref{fig:DirectModelSimulation}), and the respective speed and slope profiles along the drive.}
    \label{fig:TimeLineResults}
 \end{figure}

Fig.~\ref{fig:TimeLineResults} shows a comparison of real and model-estimated accelerations as well as the speed and slope profiles along the drive. Note that a wide range of speeds have been used, up to \unit{105}{\kilo\meter\per\hour}, covering urban and highway areas. Both accelerations are quite similar, except for some disturbances due to irregularities on the road at some points, like bumpers, or sharp turns, where the assumption of decoupled lateral and longitudinal dynamics is not satisfied. However, these situations occur over short time intervals and can be neglected in general. Also note that most acceleration measurements during the drive fall within the range which is considered comfortable ($\lvert a \rvert<$\unit{2.5}{\metrepersquaresecond}).
 
\begin{figure}[!t]
    \centering
    \includegraphics[width=1.0\linewidth]{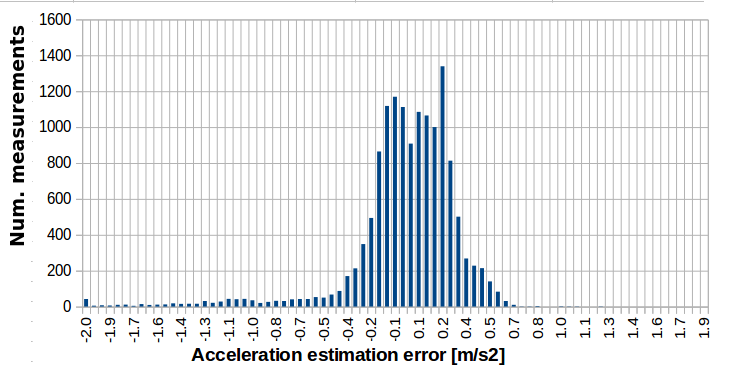}
    \caption{Histogram of the error in acceleration estimation}
    \label{fig:AccelErrorHistogram}
 \end{figure}

 Table ~\ref{tab:ErrorStatisticsTable} and Fig.~\ref{fig:AccelErrorHistogram} show the statistics of the error of the acceleration estimated by the model with respect to the real acceleration of the car. The errors are well centered, with a standard deviation of \unit{0.35}{\metrepersquaresecond}, which is less than $10\%$ of the acceleration range. 
 
\begin{table}
  \center
  \caption{Error statistics summary}
  \begin{tabular}{|l|c|}
     \hline
      & real acc. - model acc. \\
     \hline
     Average [\metrepersquaresecond] & -0.01 \\
     \hline
     Std. dev [\metrepersquaresecond] & 0.35 \\
     \hline
     Min [\metrepersquaresecond] & -2.52 \\
     \hline
     Max [\metrepersquaresecond] & 1.23 \\
     \hline
     Number of measurements & 14111 \\
     \hline
     Total distance & \unit{9.84}{\kilo\meter} \\
     \hline
   \end{tabular}
   \label{tab:ErrorStatisticsTable}
 \end{table}
 
The accuracy of the model at low speed is not ascertained by the above tests. They are indirectly validated by the fact that the model considerably improves the smoothness of speed control during slow maneuvers.

\addtolength{\textheight}{-12cm}   


\section{Conclusion}
In this paper, we propose a simple and straightforward methodology to obtain a realistic and precise model for the friction, propulsion and braking forces of an electric car for later use in speed control as an autonomous vehicle and also for the simulation of its dynamic behavior. The model obtained uses approximation functions instead of parameters for characterization. For both propulsion and brake models, we build the direct model that returns the force as a function of the input variables: speed and throttle for the propulsion model, and speed and brake for the brake model, and also the inverse model that returns the throttle as a function of speed and propulsion force, and brake signal as a function of the speed and braking force.
One of the advantages of our methodology is that they include some of the functionalities already implemented in the car like \textit{ABS} and regenerative battery charging, which are non-linear systems difficult to model using parameters. The models obtained for the propulsion and braking forces are characteristic of the vehicle itself and do not depend on external environmental variables. However, the friction may vary as a function of wind speed, tire pressure and road characteristics. When the models are used for speed control, the controller is in charge of compensating for these variations between reality and the model, so that the friction model obtained can still be used despite these differences. 
In the experimental section, we validate the model proposed by comparing the real acceleration with the model-estimated along a drive that includes a wide range of speeds, road slope changes and also a range of accelerations and decelerations that enter within the limits of comfort on a normal trip. 


\section*{Acknowledgements}
The authors would like to thank the HIANIC project consortium for its collaboration in this paper. We would also like to thank the Département de Loire-Atlantique for providing access to the race track "Circuit de Loire-Atlantique", which was fundamental for this work.


\bibliographystyle{IEEEtran}
\bibliography{biblio}

\end{document}